%% file: sn-article.tex
\documentclass[pdflatex,sn-mathphys-num]{sn-jnl}% Math and Physical Sciences Numbered Reference Style
\usepackage{graphicx}%
\usepackage{multirow}%
\usepackage{amsmath,amssymb,amsfonts}%
\usepackage{amsthm}%
\usepackage{mathrsfs}%
\usepackage[title]{appendix}%
\usepackage{xcolor}%
\usepackage{textcomp}%
\usepackage{manyfoot}%
\usepackage{booktabs}%
\usepackage{mathtools}%
\usepackage{listings}%
\usepackage[ruled,vlined]{algorithm2e}

\theoremstyle{thmstyleone}%
\theoremstyle{thmstyletwo}%

\theoremstyle{thmstylethree}%

\begin{document}

\title[Trajectory-Based Difficulty Score]{Trajectory-Based Difficulty Scoring for Reliable Learning on Tabular Data}


\author*[1]{\fnm{Tomer} \sur{Lavi}}\email{tomerlav@post.bgu.ac.il}
\author{\fnm{Bracha} \sur{Shapira}}\email{bshapira@bgu.ac.il}
\author{\fnm{Nadav} \sur{Rappoport}}\email{nadavrap@bgu.ac.il}

\affil*[1]{\orgdiv{Faculty of Computer and Information Science}, \orgname{Ben-Gurion University of the Negev}, \orgaddress{\country{Israel}}}


\abstract{
Gradient-boosted trees achieve strong performance on tabular data, yet often leave a long tail of poorly predicted instances. We introduce a Trajectory-based Difficulty Score (TDS), an instance-level difficulty estimator for boosted ensembles derived from per-tree cumulative prediction trajectories. For each instance, we compute interpretable trajectory descriptors (e.g., variance, oscillation peaks, sign switches, and tail stability) and train a lightweight regression model to predict held-out loss. An empirical CDF calibrates the resulting signal into a score in $[0,1]$ that supports ranking hard cases. Across diverse tabular benchmarks and ensemble sizes, TDS exhibits strong rank correlation with error and outperforms established instance-hardness and uncertainty baselines on classification, while remaining competitive on regression. We then show how a single difficulty signal improves multiple data mining workflows: difficulty-driven active learning for label-efficient training, difficulty-thresholded selective prediction for improved risk-coverage trade-offs, and TDS-stratified (Mondrian) conformal prediction for more uniform conditional coverage. Finally, clustering high-TDS instances using SHAP attributions reveals coherent failure modes characterized by compact feature-value ranges, supporting error analysis and targeted data acquisition.
}




\maketitle

\section{Introduction}
Tabular data is prevalent across numerous domains, making it a primary modality in many disciplines \cite{shwartz2022tabular}. It is widespread in practical applications: spanning from electronic health records to census statistics, from cyber-security to credit evaluations, and from financial industries to the natural sciences. Despite its extensive presence, it receives minimal research focus and significantly trails in both scale and capability. In contrast to the burgeoning studies on LLMs, the exploration of tabular data remains largely overlooked \cite{van2024tabular}.

Tabular data remains the stronghold of gradient-boosting models. While deep nets dominate vision and language, boosted trees (e.g., XGBoost \cite{chen2016xgboost}, Gradient Boosting Machines - GBMs \cite{friedman2001greedy}) set the state-of-the-art on structured data. Still, even state-of-the-art boosted trees leave a “long tail” of poorly predicted rows. Many real-world deployments are judged by worst-case or per-segment error, i.e. weak samples/segments, so boosting overall accuracy on those difficult instances is critical.

Existing research on sample difficulty has largely focused on confidence or loss-based indicators \citep{jiang2019fantastic, malinin2018predictive}. Measures such as predicted probability, entropy, or margin magnitude indicate uncertainty but ignore the ensemble output sequence during inference. Other works have examined gradient norms or example-forgetting statistics to estimate difficulty, yet these methods are typically defined for neural networks and do not exploit the sequential structure of boosting ensembles \cite{agarwal2022estimating, toneva2018empirical}. Consequently, the fine-grained dynamics through which individual samples are processed by successive trees remain under-explored in the literature.

In boosted-tree ensembles, each tree adds an incremental margin or log-odds such that the sequence across trees forms a trajectory. Large oscillations, sign switches and high variance in that trajectory are strong indicators of future error. By ranking samples with trajectory statistics, we can:
\begin{itemize}
    \item Emphasize difficult samples during training via re-weighting, fine-tuning, or curriculum scheduling.
    \item Drive active learning loops by iteratively querying the most difficult samples from pools of unlabeled, abandoned, or synthetic data.
\end{itemize}

\begin{figure}
    \centering
    \includegraphics[width=1\linewidth]{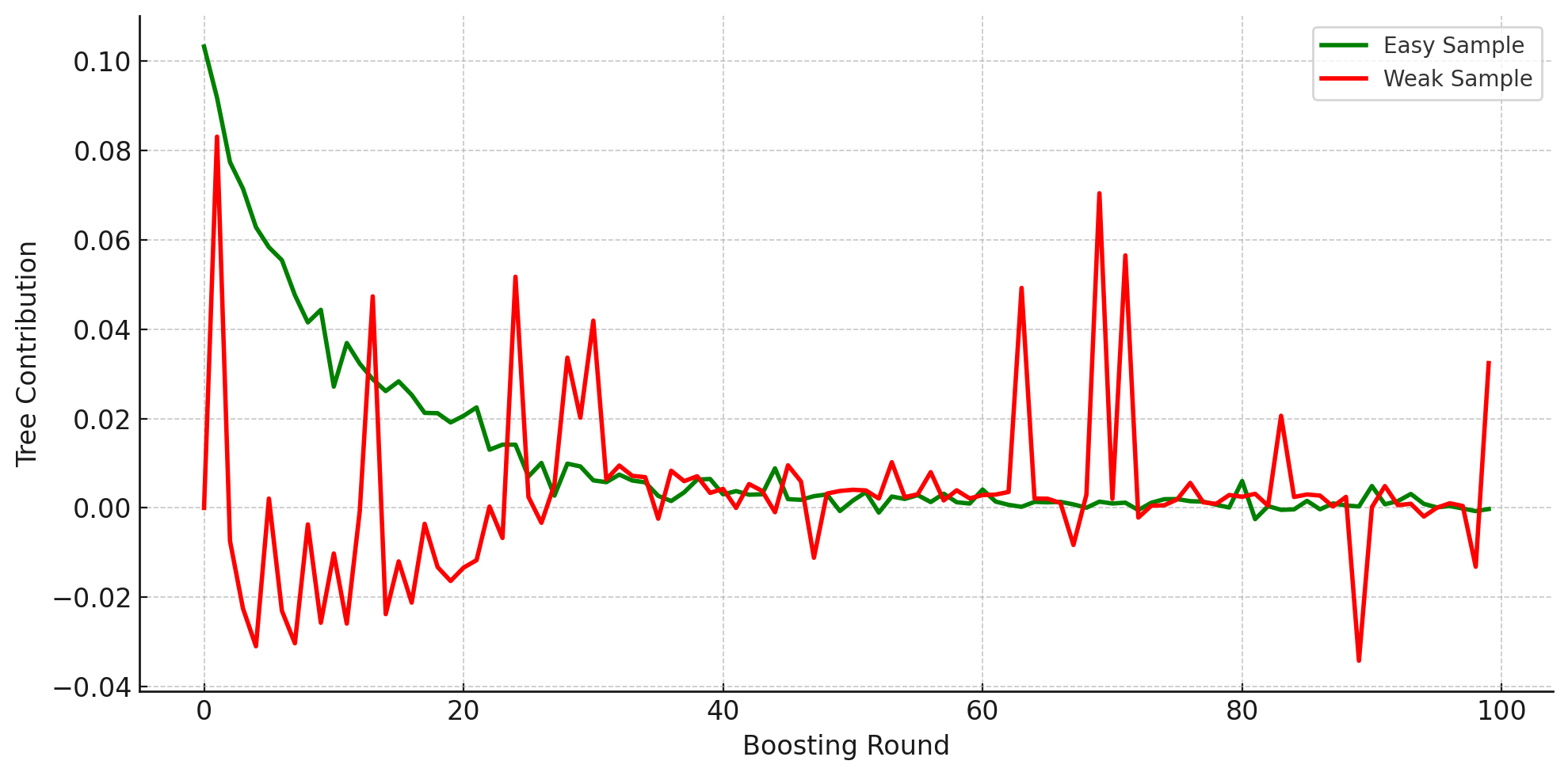}
    \caption{Comparison of tree contribution trajectories of easy (green) and weak/difficult (red) samples. The green trajectory is almost monotonic, indicating a smooth convergence to the predicted output, while the fluctuations of the red trajectory indicate the weak reliability of the model on this sample and its likelihood to error on it.}
    \label{fig:Trajectories Comparison}
\end{figure}

Figure \ref{fig:Trajectories Comparison} illustrates the trajectory attributes we aim to quantify. From the early oscillations, through the overall appearance of the curves to the late oscillations of each trajectory. The figure compares between easy and weak (difficult) samples trajectories. While the trajectory of the easy samples shows almost monotonic convergence, the weak sample is characterized by a fluctuant trajectory that doesn't fully converge throughout the inference, yielding significant tree contribution up to the end of the ensemble.
Addressing these poorly predicted instances requires not only improving the model's overall fit but also developing targeted strategies to identify and prioritize the most challenging samples during training and evaluation.

This paper proposes a Trajectory-based Difficulty Score (TDS) that directly models the internal dynamics of boosted tree-ensemble methods. Instead of relying on a single-point confidence measure, TDS constructs for each sample a trajectory capturing the output contribution of every tree in the ensemble. These trajectories present information about oscillation patterns that reflect how a model converges on each instance. From these trajectories, we extract interpretable features such as variance, peak amplitude, area under the curve, sign switches and tail stability that describe the model’s behavior on each sample. A regression model coupled with an Empirical Cumulative Density Function (ECDF) fuses these features into a difficulty score that strongly correlates with empirical prediction error. 

Beyond characterization, TDS enables several applications:
\begin{enumerate}
    \item Active learning: TDS scores serve as an acquisition function for sample selection, guiding labeling or re-sampling toward challenging regions of the input space.
    \item Selective and conformal prediction: conditioning risk-coverage trade-offs or prediction-set calibration on TDS improves reliability, especially for sub-populations prone to mis-coverage.
    \item Explainability and data enrichment: clustering high-difficulty samples by their SHAP values reveals sub-groups with typical feature-value ranges, which can promote interpretable sample acquisition in active learning and explainable model output on test samples.
\end{enumerate}

We evaluate TDS-driven methods across diverse benchmark datasets for both regression and classification tasks, comparing it against established difficulty / uncertainty measures and downstream application methods. Experiments show that trajectory-based features achieve higher correlation with true error than baseline indicators, and that TDS-driven selection strategies consistently improve active learning efficiency and conformal-prediction coverage balance. These findings highlight the potential of trajectory analysis as a general tool for diagnosing and enhancing ensemble models.

In summary, the main contributions of this study are threefold:
\begin{enumerate}
    \item A novel trajectory-based method to quantify sample difficulty in gradient boosting ensembles, derived from interpretable oscillation and convergence features.
    \item A unified approach that fuses trajectory and explainability signals to support multiple downstream tasks, including active learning, selective prediction, conformal prediction.
    \item Comprehensive empirical validation on tabular datasets, demonstrating strong correlations with error, improved performance in data-efficient active learning, and enhanced interpretability of weak samples behavior.
\end{enumerate}

\section{Related Work}

\subsection{Instance-hardness and data complexity}
Work on instance-hardness metrics, such as k-Disagreeing Neighbors \cite{smith2014instance} and Cook’s Distance \cite{cook1977detection} focus on measuring data complexity from a purely geometric perspective. 
Dynamic Instance Hardness \cite{zhou2020curriculum} and Class-Overlap Margin \cite{ho2002complexity},  Quantile-Forest Prediction Interval Width \cite{meinshausen2006quantile} and Random-Forest predictive variance \cite{breiman2001random} measure sample difficulty based on model uncertainty on a calibration set.
These approaches characterize how samples relate to their neighbors or class boundaries in feature space and model output, but they do not account for the internal dynamics of a trained model.

\subsection{Ensemble Dynamics and Trajectory Analysis}
Gradient boosting has been analyzed extensively from an optimization perspective by formalizing boosting as functional gradient descent \cite{friedman2001greedy}. Theoretical analyses have studied the relationship between large positive margins and generalization performance  \citep{brown2005diversity, rudin2009margin}, where the margin $y \cdot f(x)$ measures the signed confidence of a classification (with $y \in \{-1,+1\}$ the true label and $f(x)$ the ensemble's output). While these works highlight model-level behavior, few examine per-sample trajectories: the sequence of partial predictions that each instance generates. A few recent efforts visualize the progression of margins or the change in residuals during boosting to interpret model confidence, though without formalizing them into difficulty metrics \citep{rudin2009margin, telgarsky2013margins}.

\subsection{Explainability and failure-mode discovery in tree ensembles}
Explainability methods have become central to understanding the decision mechanisms of tree ensembles. SHAP values (SHapley Additive exPlanations) are used to explain the output of a machine learning model by attributing the contribution of each feature to the model's predictions \cite{lundberg2017unified}. TreeSHAP extends this framework efficiently to gradient-boosting models \cite{lundberg2020local}. Beyond per-sample interpretation, SHAP representations have been applied to cluster analysis and error diagnosis, revealing that groups of samples with similar attribution patterns often correspond to coherent sub-populations or failure modes \cite{covert2020understanding}. Several works have explored the integration of such explanatory signals with predictive modeling to enhance reliability and transparency \cite{arrieta2020explainable}.

\subsection{Downstream Applications of Difficulty Measures}
Difficulty and uncertainty estimates are pivotal in many downstream tasks:
\begin{itemize}
    \item In Active Learning (AL), the learning algorithm selectively queries the most informative instances for training, thereby reducing the number of labeled samples required to achieve a desired level of performance \cite{settles2009active}. This is achieved through a query mechanism: a strategy that ranks available samples according to a scoring function and selects those expected to provide the greatest benefit to the model. The effectiveness of AL critically depends on the quality of this scoring function. For scenarios where difficult or high-error samples are the most valuable for model improvement, a reliable difficulty score is essential. Such a score enables the query mechanism to focus on the hardest cases, ensuring that labeling and computation are concentrated where they can yield the greatest performance gains.
    \item In selective prediction, confidence scores guide dynamic abstention to achieve desired risk-coverage trade-offs \cite{geifman2017selective}.
    \item Conformal prediction (CP) turns point into set and value interval predictions with user-specified coverage level \(1-\alpha\) and finite-sample, distribution-free guarantees under exchangeability \citep{vovk2005algorithmic,angelopoulos2021gentle}. In CP, residual quantiles or non-conformity scores calibrate predictive sets and value intervals.
\end{itemize}
These applications demonstrate how measures of sample difficulty influence both learning efficiency and predictive reliability.

In downstream applications, we position TDS not as a plug-in to existing heuristics but as the underlying mechanism of a new method family. For active learning, TDS serves as an acquisition rule, which we compare head-to-head against a variety of baseline methods. For selective prediction and conformal prediction, we define acceptance and conditioning strategies based on TDS and evaluate them against parallel baselines. The task formulations remain standard, what differs is the decision mechanism that drives selection, acceptance, or conditioning. Accordingly, Section \ref{sec:experiments} presents empirical comparisons in difficulty scoring, AL, selective prediction and CP settings, showing how TDS enhances performance and calibration without altering the underlying task formulation.

\subsection{Gap and Contribution}
\subsubsection{Research gap}
Existing difficulty estimates for tabular models mostly express uncertainty as static, single-point signals (e.g., confidence, entropy, margin) or rely on training-time based features e.g., gradient norms and forgetting events as are defined for neural networks. For gradient boosting ensembles, these proxies overlook the inference-time oscillations of a sample’s prediction across trees that indicate whether the prediction is consistent across the ensemble. Moreover, downstream methods (active learning, selective prediction, conformal prediction) did not explore a trajectory difficulty based mechanism grounded in the behavior of tree ensembles. Finally, there is no integrated pipeline that (i) ranks samples by trajectory dynamics, (ii) clusters weak samples to reveal sub-populations, and (iii) leverages these for active learning interpretability and test-time explanation.

\subsubsection{Contributions}
This work introduces a trajectory-based method for difficult sample analysis in gradient boosting and instantiates a new family of downstream methods based on trajectory difficulty:
\begin{enumerate}
    \item Trajectory-based Difficulty Score (TDS). We define per-sample inference trajectories from cumulative tree contributions and engineer interpretable features (variance, peaks, area under trajectory, sign switches, tail stability). A lightweight fusion maps these into a continuous difficulty score that correlates with error and surfaces weak samples for both regression and classification.
    \item A new decision mechanism for downstream tasks. We instantiate trajectory difficulty based methods:
    \begin{itemize}
        \item An active learning acquisition rule that selects batches by TDS.
        \item A selective prediction acceptance rule that traces risk-coverage curves using TDS thresholds.
        \item Conformal prediction strategies that use TDS for post-CP abstention and for Mondrian (stratified) conditioning via TDS bins, keeping standard CP guarantees and pipelines. 
    \end{itemize}
    Across tasks, we compete with canonical families under identical protocols.
        \item Cluster-aware analysis. We cluster high-TDS samples to expose sub-populations with characteristic feature ranges, enabling: (i) targeted active learning and (ii) case-level explanations that pair trajectory signals with feature attributions at test time.
    \item Comprehensive evaluation on tabular benchmarks.
We report: (i) correlation of TDS with error, (ii) data-efficiency gains in active learning (AULC, best-by-val/test), (iii) improved risk-coverage in selection (AURC/NAURC), (iv) conditional coverage/width behavior in conformal prediction and (v) ablations isolating trajectory features.
\end{enumerate}

Together, these contributions establish trajectory difficulty as a principled mechanism for improving performance, reliability, interpretability and explainability of gradient boosting systems.

\section{Methodology}
\label{sec:method}
We introduce a trajectory-based difficulty estimator (TDS) for boosted-tree models on tabular data. The method has three stages: (i) constructing per-sample prediction trajectories, (ii) extracting oscillation-trend features from each trajectory, and (iii) learning a calibrated scalar difficulty score in $[0,1]$ that is reusable across tasks. Algorithm \ref{alg:tds} describes the full procedure.

\subsection{Trajectory Construction}
To quantify sample difficulty in gradient-boosted decision tree models, we begin by extracting prediction trajectories for each individual sample. These trajectories are sequences of intermediate model outputs produced at each boosting round. Specifically, for a sample $x_i$, we record the raw output of the ensemble after each of the first $T$ trees:
\[ \mathbf{t}_i = [F_1(x_i), F_2(x_i), \dots, F_T(x_i)] \in \mathbb{R}^T \]
where $F_t(x_i)$ is the cumulative prediction of the $t$-th tree.
For classification tasks, these outputs are the cumulative log-odds (margins), which can be converted to predicted probabilities via the logistic sigmoid function. For regression tasks, they represent intermediate point estimates. For each sample, we compute two types of trajectories: the sequence of model prediction oscillations (raw log-odds in classification, predicted values in regression), and the residual trajectory defined as the prediction loss trends at each iteration. We combine both oscillation and trend trajectories, enabling dual-mode analysis of model behavior over boosting rounds.

\subsection{Trajectory Features}
From the trajectory $\mathbf{t}_i$, we extract a vector of statistical features that characterize its dynamic behavior over the boosting rounds. These features are designed to capture various forms of instability in the model's prediction process, which we hypothesize to correlate with sample difficulty. The feature set includes:
\begin{itemize}
    \item Standard Deviation (STD) and Median Absolute Deviation (MAD).
    \item Trajectory peak magnitude (MAD from initial value).
    \item Area Under the Curve (AUC) of $|\mathbf{t}_i|$ and Total Area for $|\mathbf{t}_i| > \delta$.
    \item Longest monotonic segment and number of sign switches.
    \item Head vs. tail slope, AUC and STD (e.g., first 10 vs. last 10 steps).
    \item Ratio features comparing early vs. late trajectory segments.
\end{itemize}
Each feature is optionally sign-adjusted to ensure a positive association with difficulty: higher values should indicate higher difficulty. This yields a fixed-length feature vector $\phi_i \in \mathbb{R}^d$ per sample. 

\subsection{Difficulty Score}
To transform trajectory features into a quantitative difficulty score, we train a meta-model that learns to predict the held-out error of the base learner. For each sample $x_i$, we compute the squared error of the model's prediction:
\[ err_i = \text{err}(x_i) = (f(x_i) - y_i)^2 \]
A regressor is then trained to map $\phi_i \mapsto err_i$. This model learns which oscillatory patterns in the trajectory correspond to eventual high error, enabling generalization to unseen samples.
The output of this regressor, denoted $d_i = \hat{y}_i$, serves as the raw difficulty score. An ECDF (Empirical Cumulative Density Function) maps the output to a calibrated difficulty score in $[0,1]$ which reflects the model's confidence in its own performance, derived entirely from internal dynamics rather than prediction confidence or label information.

\subsection{Complexity}
Extracting cumulative trajectories for $n$ instances and $T$ trees is $O(nT)$ time. Feature extraction is $O(nT)$ for single-pass descriptors (variance, sign switches, segmented statistics). Training the difficulty regressor is $O(nd \cdot C_g)$ where $d$ is the number of trajectory features and $C_g$ depends on the regressor family. Memory can be $O(nT)$ if trajectories are stored, or $O(n)$ if features are streamed per instance.

\begin{algorithm}[ht]
\caption{Trajectory-based Difficulty Score (TDS) for boosted-tree ensembles (cumulative outputs)}
\label{alg:tds}
\KwIn{
A trained boosted-tree ensemble $M=\{f_1,\dots,f_T\}$;\\
An evaluation set $\mathcal{D}_{\mathrm{eval}}=\{(x_i,y_i)\}_{i=1}^{n}$;\\
A loss function $\ell(\hat{y},y)$ (e.g., squared error, log-loss).
}
\KwOut{
A scoring function $\mathrm{TDS}(x)\in[0,1]$ that assigns higher scores to more difficult samples.
}

\BlankLine
\textbf{Definitions.}
Let $F_t(x)$ be the \emph{cumulative} ensemble output after $t$ trees:
\[
F_0(x)=0,\quad F_t(x)=F_{t-1}(x)+f_t(x)\;\;\;\text{for }t=1,\dots,T.
\]
The cumulative trajectory of $x$ is $\mathbf{t}(x) = [F_1(x),\dots,F_T(x)] \in \mathbb{R}^{T}$.

\BlankLine
\textbf{Fit the difficulty regressor on $\mathcal{D}_{\mathrm{eval}}$.}\\
\For{$i \leftarrow 1$ \KwTo $n$}{
  $\mathbf{t}_i \leftarrow \mathbf{t}(x_i) = [F_1(x_i),\dots,F_T(x_i)]$\;
  $\boldsymbol{\phi}_i \leftarrow \mathrm{Feat}(\mathbf{t}_i)$ \tcp*[r]{trajectory features}
  $\hat{y}_i \leftarrow F_T(x_i)$ \tcp*[r]{final cumulative output}
  $err_i \leftarrow \ell(\hat{y}_i, y_i)$ \tcp*[r]{sample error / loss}
}
Fit a regression model $g$ on $\{(\boldsymbol{\phi}_i, err_i)\}_{i=1}^{n}$\;
\For{$i \leftarrow 1$ \KwTo $n$}{
  $\hat{d}_i \leftarrow g(\boldsymbol{\phi}_i)$ \tcp*[r]{predicted difficulty (unnormalized)}
}
Fit an empirical CDF $\widehat{C}$ over $\{\hat{d}_i\}_{i=1}^{n}$:
\[
\widehat{C}(u)=\frac{1}{n}\sum_{i=1}^{n}\mathbb{I}\{\hat{d}_i \le u\}.
\]

\BlankLine
\textbf{Define the TDS scoring function for any input $x$.}\\
\SetKwFunction{TDS}{TDS}
\SetKwProg{Fn}{Function}{:}{}
\Fn{\TDS{$x$}}{
  $\mathbf{t} \leftarrow [F_1(x),\dots,F_T(x)]$\;
  $\boldsymbol{\phi} \leftarrow \mathrm{Feat}(\mathbf{t})$\;
  $\hat{d} \leftarrow g(\boldsymbol{\phi})$\;
  \Return $\widehat{C}(\hat{d})$\tcp*[r]{$\in[0,1]$; higher means more difficult}
}
\end{algorithm}

\subsection{Theoretical Foundations of Trajectory-Based Difficulty}
The strong empirical correlation between trajectory dynamics and prediction difficulty is grounded in gradient boosting's fundamental properties. We provide theoretical insight into why trajectory features capture sample difficulty, under what conditions the method succeeds or fails, and how TDS relates to established boosting theory.

\subsubsection{Why Trajectory Dynamics Correlate with Prediction Difficulty}
Gradient boosting operates as functional gradient descent, where each tree minimizes the residual error of the preceding ensemble \cite{friedman2001greedy}. For a sample $x_i$, the trajectory $\mathbf{t}_i$ represents cumulative predictions across boosting rounds. When the underlying feature-target relationship is well-captured by the model, residuals \(r_t(x_i) = y_i - \hat{y}_t(x_i)\) decrease steadily, producing smooth, converging trajectories. Conversely, trajectories of difficult samples fail to converge smoothly, with trees repeatedly adjusting predictions in conflicting directions. Large oscillations, sign switches, and high variance in trajectories directly indicate prediction uncertainty. This instability captures the model's difficulty to confidently learn the mapping for that sample. In classification, trajectory volatility directly reflects margin dynamics: samples with oscillating trajectories exhibit smaller or fluctuating margins across rounds, providing a dynamic view of margin evolution that static measures cannot capture.

\subsubsection{Conditions for Method Success and Failure}
\paragraph{Success}
TDS performs well when difficult samples arise from learnable complexity (class overlap, non-linear relationships). The ensemble learns convergence patterns when labels are clean or systematically corrupted.

\paragraph{Failure}
TDS struggles with randomly mislabeled data. For a randomly mislabeled sample with corrupted label \(\tilde{y}_i \neq y_i\), the residual trajectory \(r_t(x_i) = \tilde{y}_i - \hat{y}_t(x_i)\) is computed from the corrupted label. If multiple nearby samples with correct labels guide the model toward the output function \(f(x) \approx y_i\), the mislabeled sample generates a trajectory moving away from \(\tilde{y}_i\). Since TDS assumes label correctness, it cannot distinguish between inherent difficulty and random label flips. Empirically, TDS struggles on datasets with known random label noise, e.g., Russian Housing Market \cite{sberbank-russian-housing-market}.

\subsubsection{Connection to Boosting Theory}
Classical boosting theory establishes that maximizing margins improves generalization \cite{rudin2009margin}. TDS complements this by providing per-sample trajectory-level margin dynamics: samples with smooth, converging trajectories achieve large, stable margins, while oscillating trajectories indicate smaller, fluctuating margins that increase generalization error. Gradient boosting minimizes loss \(\mathcal{L}(y, f(x))\) through sequential negative gradient approximation. Persistent non-zero trajectory variance and oscillations indicate that functional gradient descent has not converged for that sample, a direct difficulty indicator.

\subsubsection{Mixed Uncertainty Capture: Epistemic and Aleatoric}
TDS captures both epistemic (under-explored regions) and aleatoric (class overlap / boundary cases) uncertainty through trajectory fluctuations. In regions with class overlap, nearby samples have different labels due to inherent ambiguity. During gradient boosting, different trees observe different subsets of these overlapping samples, pushing predictions toward different classes, causing trajectory oscillations that reflect aleatoric uncertainty. TDS also captures epistemic uncertainty from insufficient data or model capacity: regions of little knowledge produce inconsistent tree corrections and unstable trajectories.

Rather than attempting to decompose TDS into separate components, which is theoretically problematic and practically challenging, we position TDS as a holistic difficulty metric capturing combined uncertainty sources as expressed through prediction dynamics. This mixed capture benefits downstream applications: active learning gains from both aleatoric and epistemic regions, selective prediction justifies abstention from either source, and conformal prediction requires wider intervals for both regions. However, it explains performance variations: in classification with significant class overlap, TDS may select redundant boundary samples which skew the label distribution. Such cases exhibit classification performance variations compared to regression. This perspective explains why TDS shows stronger, more consistent performance on regression tasks where aleatoric uncertainty from overlapping target values is less pronounced than in discrete classification.

\subsection{Downstream tasks}
We show utilization of TDS across three decision contexts. (See also empirical results in section \ref{sec:experiments}).

Given $\mathrm{TDS}:\mathcal{X}\to[0,1]$:

\subsubsection{Active Learning} 
AL selects a batch $B$ of size $b$ from the pool $\mathcal{U}_t$ by maximizing total difficulty.
\begin{equation}
\mathrm{S}_t=\arg\max_{\substack{B \subset \mathcal{U}_t}} \sum_{x\in B}\mathrm{TDS}(x)
\tag{AL}\label{eq:al}
\end{equation}
\subsubsection{Selective Prediction}
For a TDS difficulty score threshold $\tau$, accept $x$ iff $\mathrm{TDS}(x)\le\tau$, 
\begin{equation}
    \mathrm{acc}_\tau(x)=\mathbf{1}\{\mathrm{TDS}(x)\le \tau\}
    \tag{SEL}
\end{equation}
yielding coverage $\mathrm{cov}(\tau)=\mathbb{E}[\mathrm{acc}_\tau(X)]$ and conditional risk.
\[
\mathrm{risk}(\tau)=\mathbb{E}\!\left[\ell(\hat{y}(X),Y)\mid \mathrm{acc}_\tau(X)=1\right]
\]

\subsubsection{TDS-Mondrian Conformal Prediction}
Stratify calibration by $\text{bin}(x)$ (TDS quantiles) to target conditional coverage $1-\alpha$. Match sample $x$ to its bin:
\begin{equation}
\begin{aligned}
\mathrm{bin}(x)&=q(\mathrm{TDS}(x))
\end{aligned}
\tag{CP}\label{eq:cp}
\end{equation}
Compute predictive interval / set  $\Gamma_\alpha(x)=\mathrm{CP}\!\left(\text{calib}\mid \mathrm{bin}(x)\right)$ to meet the coverage statement: $
\mathbb{P}\!\left\{Y\in \Gamma_\alpha(X)\mid \mathrm{bin}(X)\right\}\approx 1-\alpha$.

\subsection{TDS-Segment (Segment-Based Active Learning)}
While difficulty-based sampling identifies individual difficult samples, it treats each sample independently. However, difficult samples often share underlying structure, representing sub-populations where the model systematically under-performs. To exploit this structure, we cluster the most difficult samples and use their common feature characteristics to guide additional sample selection.

\subsubsection{Inputs}
\begin{itemize}
    \item Unlabeled pool $\mathcal{U}$ and a calibration set $\mathcal{C}$.
    \item Segmentation $g$ with feature-range descriptors for each segment $s\in\{1,\dots,K\}$.
    \item Batch size $b$ and calibration coverage target $M\%$.
\end{itemize}

\subsubsection{Preprocessing}
\begin{enumerate}
\item For each segment $s$, compute a calibration difficulty statistic:
    \[
        D(s) \;=\; \frac{1}{|\mathcal{C}_s|} \sum_{x\in \mathcal{C}_s} \mathrm{TDS}(x)
    \]
    \item Rank segments by difficulty and choose the smallest set covering $M\%$ of $\mathcal{C}$.    
    \item Sort the segments by $D(s)$ in descending order and take the smallest prefix $\mathcal{S}^\star$ such that the calibration coverage is:
    \[
        \frac{\left| \bigcup_{s\in \mathcal{S}^\star} \mathcal{C}_s \right|}{|\mathcal{C}|} \;\ge\; \frac{M}{100}
    \]
    \item Characterize segments by feature ranges :
    \begin{itemize}
        \item For a continuous feature $j$: an interval $[a^{(s)}_j,b^{(s)}_j]$.
        \item For a categorical feature $j$: an allowed set $V^{(s)}_j$.
    \end{itemize}
\end{enumerate}

\subsubsection{Per-round procedure}
Sample the pool by range membership and return top-$b$ by TDS. Form the candidate subset of the pool
    \[
        \mathcal{U}^\star \;=\; \{\, x\in \mathcal{U} : \exists s\in \mathcal{S}^\star \text{ with } x \in s \,\}.
    \]
    Rank $\mathcal{U}^\star$ by $\mathrm{TDS}(x)$ (highest first) and return the top $b$ points.

\subsubsection{Training interpretability and Human-in-the-Loop}
Because cluster rules are expressed as value ranges over a small number of features, they can be presented to human experts for interpretation and feedback. This offers a pathway to combine automated discovery of model weaknesses with domain-informed data acquisition strategies.

\subsection{TDS as a Unified Difficulty Signal}
TDS distills model-internal dynamics into a single calibrated instance-level score. This score can be reused as a common primitive in data mining pipelines: (i) prioritizing instances for labeling or enrichment, (ii) ranking instances for abstention to control risk-coverage behavior, and (iii) stratifying calibration to reduce conditional miscoverage. Importantly, the base learner and the downstream task definitions remain unchanged and only the instance-ranking/conditioning signal is replaced.

\section{Experiments and Results}
\label{sec:experiments}

\subsection{Reproducibility}
To evaluate our method performance and robustness we performed a wide range of experiments comparing our method against strong baseline methods on a suite of tabular benchmarks (see table \ref{tab:datasets}) and a range of model architectures. 
\begin{itemize}
    \item Tasks: regression and binary classification.
    \item Model: XGBoost (classifier/regressor) with fixed grids:  n\_estimators (100/300/1000), max\_depth (6/8), learning\_rate (0.1), subsample=0.8, colsample\_bytree=0.8. For classification datasets, Temperature Scaling calibration \cite{guo2017calibration} was applied with optimizer='BFGS' and maxiter=50.
    \item Splits: ten repeated holdouts with fixed seeds:
    \begin{itemize}
        \item 60\% for training, 20\% for calibration and 20\% as a held-out test set.
    \end{itemize}
    \begin{itemize}
        \item For AL 20\% initially labeled for the first training, 60\% as the abandoned pool for acquisition during a total of 30 training rounds, 10\% for calibration and 10\% as a held-out test set.
        \item All reported result values are means with 95\% CI over datasets and model architectures.
    \end{itemize}
    \item Preprocessing:
    \begin{itemize}
        \item Fill missing values using mean values.
        \item Encode categorical features.
        \item All features are standardized.
    \end{itemize}
    \begin{itemize}
    \item Dataset specific preprocessing:
    \begin{itemize}
        \item Russian Housing: Replace the timestamp feature by month and week values.
    \end{itemize}
        \item No additional task-specific feature engineering is used beyond the trajectory construction as described in the Methodology section \ref{sec:method}.
    \end{itemize}
    \item Trajectory Construction: We extract features from the raw predictions trajectory. Optionally, according to a configurable flag, label\_free=use\_labels\_for\_trends, we add a second trends trajectory of the prediction residuals. Default label\_free=use\_residual\_trajectory for all experiments except regression AL.
    \item TDS: For our method, trajectory-difficulty features were extracted using the fixed hyperparameters \(\delta=0.1\), \(\text{head\_size}=0.3\), \(\text{tail\_size}=0.2\), and the segment-based variant used \(n_{\text{clusters}}=4\) with the top \(n_{\text{fi}}=5\) features per cluster to define value-range rules.
    \item Compute: All experiments were performed on a Macbook Pro M4 Max 36GB RAM. 
    \item Code for reproducing our experiments and summary results are available at \url{https://anonymous.4open.science/r/TDS-1282}.

\end{itemize}

\begin{table}[ht]
  \centering
  \label{tab:datasets}

\input{Tables/table_datasets}
\end{table}

\subsection{Difficulty Scores}
\subsubsection{Baseline Methods}
We assess whether sample difficulty scores correlate with the model's actual error by computing Pearson's $r$ \cite{pearson1895vii} and Spearman's $\rho$ \cite{spearman1904proof} between each method's scores and prediction error on held-out data (per dataset), then averaging across datasets. For TDS we prioritize Spearman's $\rho$ which emphasizes ranking of samples and thus is appropriate for identifying the most difficult samples. 
Baselines used for this experiment are: Cook’s Distance, Quantile-Forest Prediction Interval Width (QRF\_PI\_width), Random-Forest predictive variance (RF\_var), class overlap, Dynamic Instance Hardness (DIH) and k-Disagreeing Neighbors (kDN).

\subsubsection{Results}
Across datasets and tree budgets, TDS shows consistent rank association with per-sample error. This pattern validates TDS as a difficulty signal: it reliably orders samples by difficulty, even when absolute calibration differs, making it well-suited as a backbone for downstream tasks where relative difficulty is the operative criterion. We therefore prioritize Spearman as a primary ranking metric for TDS optimization.
For the binary classification datasets TDS Spearman correlation outperforms all baselines, scoring $\rho$ $\approx$ 0.75-0.96. TDS scores Pearson $r$ $\approx$ 0.16-0.57. On the regression datasets average Spearman $\rho$ is strong in the regime most relevant to our pipelines ($\rho$ $\approx$ 0.39-0.43), rank typically 3/5 and close to the top method, while Pearson $r$ is smaller ($r$ $\approx$ 0.11-0.24), as expected for a score designed for monotonic ranking rather than linear fit. 

\paragraph{Statistical significance}
Across the classification datasets, a paired Wilcoxon signed-rank test on Spearman difficulty error correlation shows that TDS significantly outperforms the best baseline (DIH) at every capacity (100, 300, and 1000 trees, one-sided $p = 0.0117, 0.0117, 0.0391$, respectively. Holm-corrected $p \le 0.039$), with positive median improvements in all cases.

Table~\ref{tab:correlations} summarizes the mean correlations (higher is better). 
\begin{table}[ht]
  \centering
  \caption{Mean correlation of difficulty scores with error across datasets (higher is better). Values are mean with 95\% CI computed across datasets.}
  \label{tab:correlations}

\input{Tables/table_difficulty_correlations}
\end{table}

\subsection{Downstream Tasks}
We evaluate TDS based implementation of three downstream tasks.

\subsubsection{Active Learning by Difficulty Ranking}
\paragraph{Active Learning Loop}
Let $\mathcal{D}_{\text{train}}^{(0)}$ denote the initial labeled dataset, $\mathcal{D}_{\text{pool}}$ the pool of unlabeled samples, and $\mathcal{M}^{(t)}$ the model trained at iteration $t$. The difficulty-based active learning process proceeds as follows:
\begin{enumerate}
    \item Train the model $\mathcal{M}^{(t)}$ on the current labeled set $\mathcal{D}_{\text{train}}^{(t)}$.
    \item Use the model to compute prediction and residual trajectories for each $x \in \mathcal{D}_{\text{pool}}$, and extract difficulty features.
    \item Fit the trajectory method on $\mathcal{M}^{(t)}$ using an evaluation set $\mathcal{D}_{\text{val}}^{(t)}$.
    \item Apply the difficulty regressor to obtain difficulty scores $d_x$ for each sample.
    \item Select the top $k$ samples with the highest difficulty scores: $\mathcal{S}^{(t)} = \arg\max_{x \in \mathcal{D}_{\text{pool}}} d_x$.
    \item Query labels for $\mathcal{S}^{(t)}$, remove them from $\mathcal{D}_{\text{pool}}$, and add them to the training set:
    \[ \mathcal{D}_{\text{train}}^{(t+1)} = \mathcal{D}_{\text{train}}^{(t)} \cup \mathcal{S}^{(t)} \]
    \item Repeat the process for a fixed number of iterations or until budget exhaustion. 
\end{enumerate}

This strategy prioritizes samples that the current model is most likely to fail, as inferred from internal signal dynamics rather than output probabilities alone.

\paragraph{Implementation Details}
\subparagraph{Trajectory Features}
We extract trajectories using configurable $\delta$, head and tail sizes. We mix weak and representative samples from the pool to avoid overfit to outliers for a configurable length of 3 first iterations on regression models and .5 mix ratio. For the classification models we mix samples throughout the entire training process. For the experiments on model size of 1000 estimators which present a long tail of model convergence, we reduce the ensemble size by selecting a configurable head of 400 trees and compressing them to a configurable size of 100 trees using block mean values. Compression to 300 trees is also done on 300 estimator models.

\paragraph{Segment-Based Active Learning:}
We begin by selecting a subset $\mathcal{D}_{\text{hard}} \subset \mathcal{D}_{\text{pool}}$ consisting of the top $p\%$ most difficult samples according to their calibrated difficulty scores $s_i$. Each sample is then represented by a vector of SHAP values reduced via PCA by a configurable size (default=5):
\[ z_i = \text{PCA}(\text{SHAP}(x_i)) \in \mathbb{R}^k \]
We then cluster these representations, forming $K$ segments:
\[ \mathcal{D}_{\text{hard}} = \bigcup_{j=1}^K C_j \quad \text{with} \quad C_j = \{ x_i \in \mathcal{D}_{\text{hard}} : \text{cluster}(x_i) = j \} \]
We then use K-means \cite{lloyd1982least} to divide the PCA vectors to a configurable partition to $K$ segments (default=4).

\paragraph{Baseline Methods}
We compare TDS and segment-based TDS to strong baselines (Random, Uncertainty variants, QBC, BADGE, Learning-Loss \cite{yoo2019learning}, Core-Set \cite{sener2017active}, EMC \cite{settles2008analysis}, GLISTER, BatchBALD). We report the Area Under the Learning Curve (AULC) \cite{settles2009active} computed over RMSE vs. labeled budget (lower is better) and the final RMSE at the end of the budget. 
AL baselines ran with the following hyperparameters:
Query By Committee: \#members=5, BADGE: standardize=true, BatchBALD: \#mc=15, dropout\_rate=.15.

\paragraph{Runtime analysis}
Table~\ref{tab:al-times} consolidates mean computation time across datasets. TDS takes a total of $\approx$ 25s to complete the 30 AL iterations. TDS-Segment requires additional $\approx$ 9s for cluster analysis.
All baseline AL methods, except BatchBALD, completed full AL in under 2s on average. BatchBALD required $\approx$ 45s due to its Bayesian entropy computation.

\begin{table}
\centering
\caption{AL average total selection time in seconds per method and ensemble size. TDS-Segment times indicate the segment analysis times that are to be added to TDS times. }
\label{tab:al-times}
\begin{tabular}{c c c}\toprule
\textbf{Method} & \textbf{Estimators} & \textbf{Avg. Sel. Time} \\\midrule
BatchBALD & 100 & 41.2 \\
 & 300 & 45.2 \\
 & 1000 & 49.7 \\
TDS & 100 & 23.9 \\
 & 300 & 25.0 \\
 & 1000 & 27.4 \\
TDS-Segment* & 100 & 8.7 \\
 & 300 & 9.1 \\
 & 1000 & 12.6 \\ \bottomrule

\end{tabular}

\end{table}

\paragraph{Results}
\subparagraph{Regression:}
On the regression task, TDS selection yields lower AULC and better final RMSE over most of the experiments using default settings with no Hyperparameter Optimization (HPO). TDS-segment also improves over the majority of the non-trajectory baselines on average. 
Table~\ref{tab:al-summary} consolidates mean performance across datasets. 

\begin{table}[ht]
  \centering
  \caption{Active learning: mean final RMSE and AULC across datasets (lower is better). Values are means with 95\% CI over datasets. Table is grouped by number of estimators in the regression model and sorted by AULC on each group.}
  \label{tab:al-summary}
  \input{Tables/table_al_summary.tex}
\end{table}

We performed HPO on the delta, head and tail sizes to improve TDS results on select experiments where it under-performed. We selected the Bike Sharing benchmark and the challenging Sberbank Russian Housing Market dataset from Kaggle competition (Russian Housing), which is known for its noisy labels. Detailed feature description is available on the competition page on Kaggle \footnote{\url{https://kaggle.com/competitions/sberbank-russian-housing-market}}.
For the Bike Sharing dataset with 1000 estimators we used delta=0.01, head\_size= 0.23, tail\_size=0.19 which improved TDS AULC from 6.750 to 6.499 and test RMSE from 5.793 to 5.695. For the Russian Housing dataset with 1000 estimators we used delta=0.04, head\_size= 0.21, tail\_size=0.096 and omitted the trends trajectory by setting label\_free=use\_residual\_trajectory, which improved TDS AULC from 0.515 to 0.487 and test RMSE from 0.494 to 0.490.
Following HPO, TDS turned to lead the results on both experiments over all the baseline methods. Yet, on max\_depth=8 we struggled to dominate the AL experiment on the  Russian Housing dataset. We conclude that TDS is sensitive to mislabeled data by random label corruption. Such idiosyncratic random label flip upon otherwise identical $x$ cannot be learned from the trajectory signals of the points.

\begin{figure}
    \centering
    \includegraphics[width=1\linewidth]{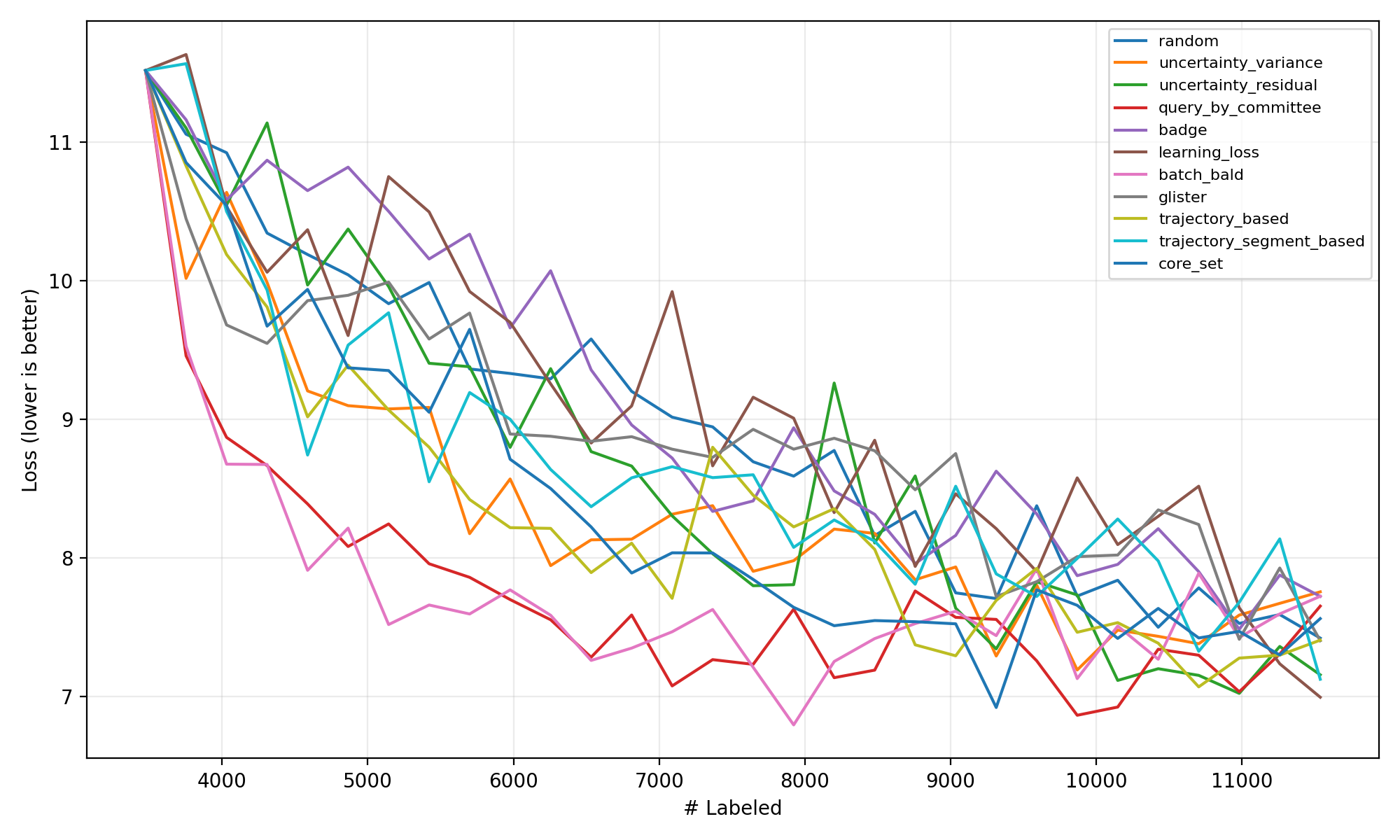}
    \includegraphics[width=1\linewidth]{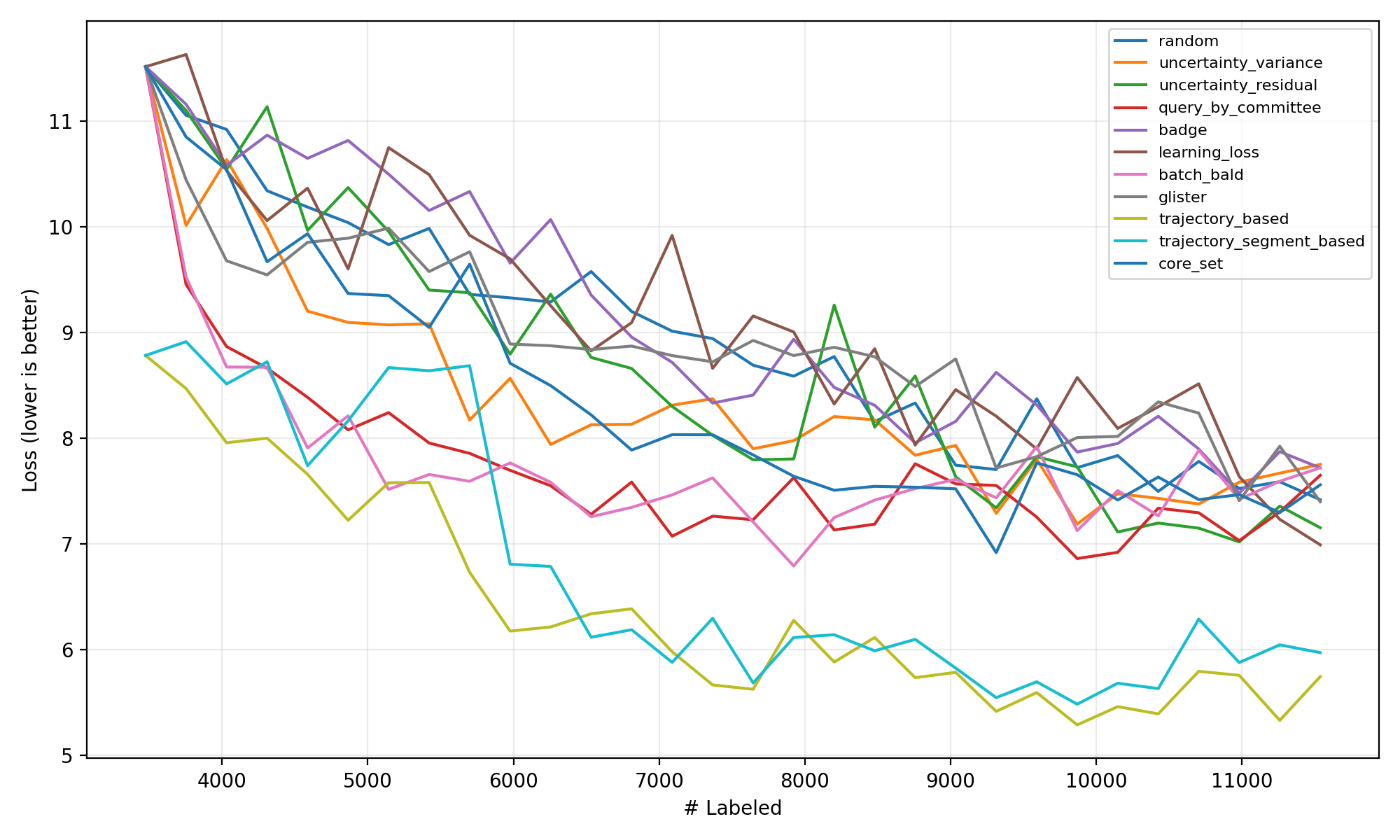}
    \caption{Learning curves of model training iterations by AL on the Bike Sharing dataset. The top figure compares baselines to un-optimized TDS and the bottom figure shows results on optimized hyper-parameters TDS.}
    \label{fig:al_bike_sharing}
\end{figure}

Figure \ref{fig:al_bike_sharing} compares the learning curves of the two Bike Sharing experiments. Using HPO, both TDS and TDS-Segment lead the results over the other baselines both in AULC and RMSE.

\subparagraph{Classification:}
Across datasets, TDS and TDS-Segment exhibit a slow warm-up but strong late-stage efficiency and reliable early-stopping performance. At all model sizes, TDS (regular) scores the lowest log-loss (LL). 
Early-phase (0-25\%) AULC: TDS exhibits a "slow start”. This is consistent with a strategy that prioritizes coverage that pays off later rather than immediate early drops. TDS (regular) ranks at the top on AULC (segment 75-100\%), reflecting excellent convergence despite a slow early phase. Crucially, at 1000 trees, TDS is top on Best-by-Val (Test) and AULC (75-100\%), indicating that its trajectory signal becomes increasingly informative in deeper ensembles, yielding superior late-curve efficiency and the most trustworthy validation-selected test performance.
TDS-Segment scores second best LL and mid-pack AULC (segment 75-100\%) on 100 trees models. At 300 trees, TDS-Segment is mid-pack AULC but last on LL. At 1000 trees TDS-Segment performance are lowest.

Table~\ref{tab:al-summary-cls} consolidates mean performance across datasets. 
\begin{table}[ht]
  \centering
  \caption{Active learning: mean final Log Loss (LL) and AULC (75-100\%) across datasets (lower is better). Values are means with 95\% CI over datasets. Table is grouped by number of estimators in the regression model and sorted by LL (Test) on each group.}
  \label{tab:al-summary-cls}
  \input{Tables/table_al_summary_cls.tex}
\end{table}

\paragraph{Explainability}
We would like now to demonstrate the explainable aspect of TDS-Segment. TDS-Segment identifies sub-populations in the data based on feature-value ranges, thus offering explainable characteristics of weak points in terms of their feature values. TDS-Segment improves trust and transparency while training by offering interpretability of the prioritization of the sample selection in each iteration and in inference by e.g. explaining why a sample is flagged to be abstained from prediction and recommended to be forwarded to further examination.

\subparagraph{Example - California House Prices:}
The California Housing dataset (Cal. Housing) includes 8 numeric, predictive attributes and the target median house value for California districts. Detailed feature description is available on the Scikit-Learn page \footnote{\url{https://scikit-learn.org/stable/datasets/real_world.html}}.
Table \ref{tab:cal-segments} shows feature ranges of the cluster which presented the largest error. According to the Lat/Long coordinates, the cluster represents district blocks in the Los Angeles area by descriptive features as detailed in the table. When this dataset is used to train a price estimate model, house features are used to explain sample selection for AL and prediction confidence when the model is used to predict house prices.

\begin{table}[ht]
    \centering
\caption{Feature value ranges of difficult samples from the California Housing dataset.}
\label{tab:cal-segments}    \begin{tabular}{ccc}
        \hline
            \textbf{Feature} & \textbf{Low} & \textbf{High} \\
        \hline
            Income& 1.675 & 4.175 \\
            Occupancy& 1.6277 & 2.4906 \\
            House Age& 16.0 & 49.0 \\
            Rooms& 2.848& 5.49\\
            Bedrooms& 1.0& 1.2\\
            Population & 391& 2223\\
            Latitude & 33.34 & 34.61 \\
            Longitude & -119.83 & -117.35 \\
        \hline
    \end{tabular}
    
\end{table}

\subparagraph{Example - ICU Survival:}
The WiDS 2020 Datathon dataset holds data from the first 24 hours of intensive care and a patient survival binary label. Detailed feature description is available on the datathon page on Kaggle \footnote{\url{https://www.kaggle.com/competitions/widsdatathon2020}}.
Table \ref{tab:wids-segments} shows feature ranges of the cluster which presented the largest error. These features include demographic, hospitalization history and medical measurements. Such detailed characteristic of difficult samples may be helpful on both data collection for model training as well as model output explainability and processing.
\begin{table}[ht]
    \centering
    \caption{Feature value ranges of difficult samples from an ICU survival dataset.}
    \label{tab:wids-segments}    \begin{tabular}{ccc}
        \hline
            \textbf{Feature} & \textbf{Low} & \textbf{High} \\
        \hline
            age & 54 & 85 \\
            pre\_icu\_los\_days & 0& 4.84\\
            apache\_4a\_icu\_death\_prob & 0.16 & 0.34 \\
            d1\_sysbp\_noninvasive\_min & 59.0 & 108.0 \\
            intubated\_apache & 0.0 & 1.0 \\
            d1\_sysbp\_min & 59.0 & 108.0 \\
            d1\_spo2\_min & 75.0 & 98.0 \\
            d1\_temp\_min & 34.9 & 36.9 \\
            ph\_apache & 7.25& 7.45\\
            elective\_surgery & 0.0 & 0.0 \\
            d1\_heartrate\_min & 52.0 & 101.0 \\
            gcs\_motor\_apache & 1.0 & 6.0 \\
        \hline
    \end{tabular}
\end{table}

\subsubsection{Selective Prediction by Difficulty Rankking}
\paragraph{Implementation Details}
We use TDS as an underlying score for selective prediction (higher = harder). For each threshold $\tau$ over TDS, accept set:
\[A_\tau={x:TDS(x)\le\tau}\]

\paragraph{Baseline Methods}
We compare TDS to baselines and report Risk-Coverage curve with risk = MSE on $A_\tau$ on regression and error rate for classification. Baselines used: regression Quantile Regression Forests (QRF), Bootstrap Ensemble Variance \cite{lakshminarayanan2017simple} and classification Max-Softmax (confidence), Margin (gap between top-2 classes), Entropy of $\hat{P}$.
Selective Prediction baselines ran with the following hyperparameters: 
QRF: n\_estimators=200, Bootstrap Ensemble Variance: n\_models=8, sample\_frac=1.0.

We compared Normalized AURC (NAURC), i.e. AURC normalized by the model’s error at full coverage so scores are comparable across datasets.

\paragraph{Results}
Table \ref{tab:sel-summary} consolidates mean performance across datasets.

\begin{figure}
    \centering
    \includegraphics[width=1\linewidth]{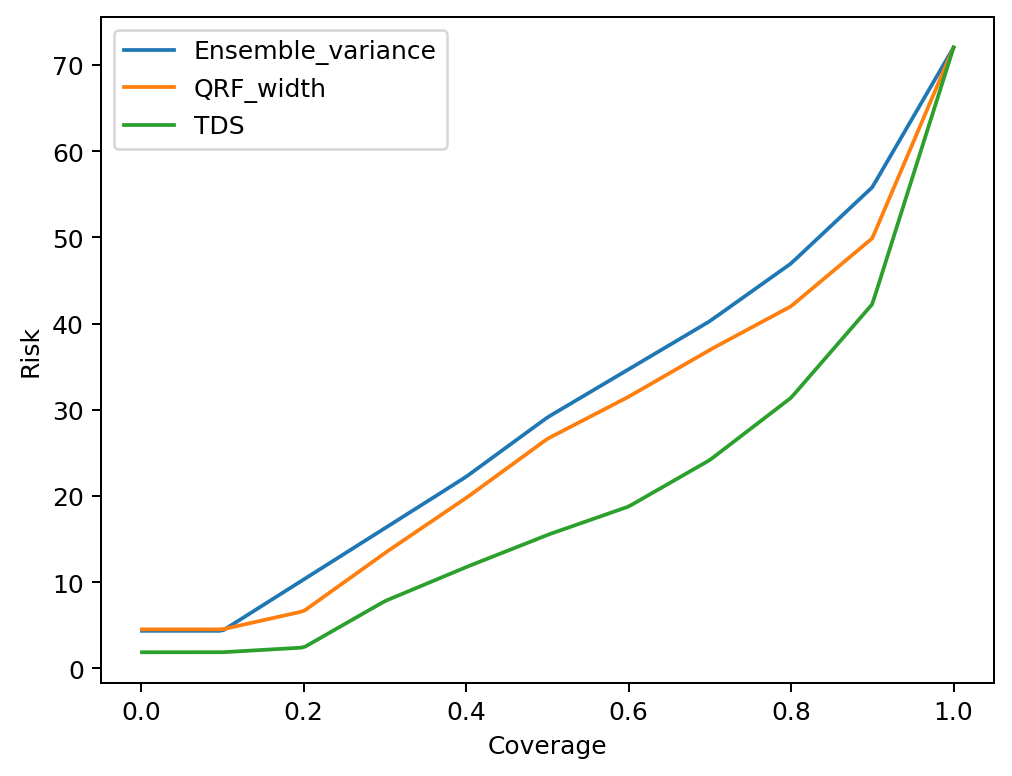}
    \caption{Risk-Coverage (MSE) curves comparison on a selective prediction task on a regression dataset. The X axis stands for the coverage decile, the Y axis is the risk of the accepted samples in MSE. TDS scores lowest risk across all deciles.}
    \label{fig:sel_bike_sharing}
\end{figure}

\begin{table}[ht]
  \centering
\caption{Selection NAURC results averaged over datasets per method and n\_estimators (lower is better). Values are means with 95\% CI over datasets. Table is grouped by
number of estimators in the model.}
  \label{tab:sel-summary}

\input{Tables/table_selection_summary}

\end{table}
\subparagraph{Regression:}
On the regression datasets, TDS’s shows comparable NAURC on 100-300 estimators and becomes the best on 1000 estimators, suggesting that stable late-boosting trajectories carry the strongest difficulty signal for ordering accept decisions.
Figure \ref{fig:sel_bike_sharing} compares the Risk-Coverage (MSE) curves of the three selective prediction methods on the Bike Sharing dataset on a regression model with 100 estimators. TDS exhibits the lowest error across all coverage thresholds.

\subparagraph{Classification:}
On the classification datasets, TDS leads on NAURC across all estimator counts, indicating stronger overall performance and better normalized ranking behavior.

\subsubsection{Conformal Prediction by Difficulty Rankking}
\paragraph{Implementation Details}
We define a TDS based conformal prediction method as follows:
\begin{itemize}
    \item    Split calibration set into (K) quantile bins of TDS.
    \item For each bin (k): nonconformity $r_i=|y_i-\hat y_i|$ (regression) or $s_i = 1-\hat{p}_i$ classification.
    \item Store the $(1-\alpha)(1+1/n_k)$ as quantile $q_k$.
    \item For test (x)
    \begin{itemize}
        \item Find its TDS bin $k(x)$.
        \item Regression output: $[\hat y(x)-q_k,\hat y(x)+q_k]$.
        \item Classification output: $\{y : \hat p_y(x) \ge 1 - q_k \}$
    \end{itemize}

\end{itemize}

\paragraph{Baseline Methods}
We compare TDS to baselines and report coverage, set size / width, Mean Abs Coverage Error (MACE), Max Coverage Error (MaxCE) and trend slope. Baselines used: Split (Vanilla) and Mondrian\cite{vovk2005algorithmic}, Conformalized Quantile Regression (CQR) \cite{romano2019conformalized} and tested for coverage $1-\alpha=0.9$.
CP baselines ran with the following hyperparameters: 
Mondrian and TDS-Mondrian: n\_bins=10, CQR base regressor parameters: n\_estimators=300, max\_depth=3, learning\_rate=0.05, min\_samples\_leaf=3.

\paragraph{Results}
Table \ref{tab:cp-summary} consolidates mean performance across datasets. For each method and ensemble size we measured coverage, width (set size for classification), MACE across TDS deciles with 1=easiest to 10=hardest, MaxCE across TDS deciles.
On both regression and classification datasets, TDS-Mondrian (TDS) consistently achieves the lowest MACE and MaxCE and has the flattest coverage trend $(|slope|\approx 0)$, i.e., it avoids under-coverage on the hardest deciles without inflating width.

\begin{table}[ht]
  \centering
\caption{Conformal Prediction results averaged over datasets per method and n\_estimators (lower is better). Values are means with 95\% CI over datasets. Table is grouped by
number of estimators in the model.}
  \label{tab:cp-summary}

\input{Tables/table_cp_summary}

\end{table}

\begin{figure}
    \centering
    \includegraphics[width=1\linewidth]{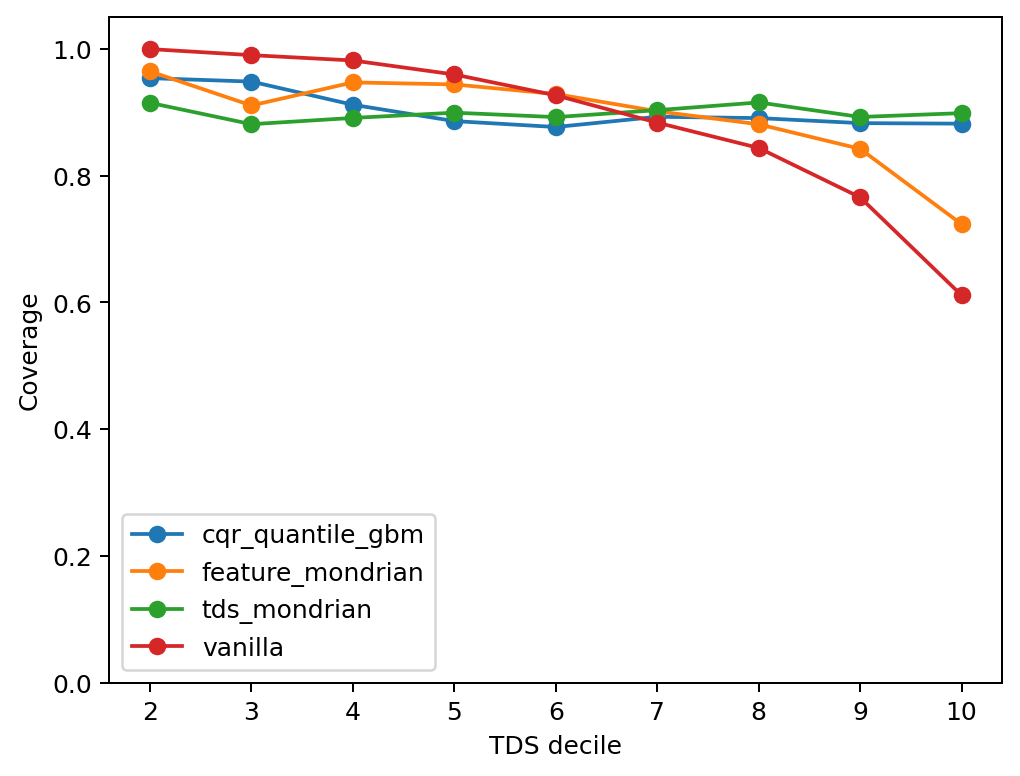}
    \caption{Coverage curves comparison on a conformal prediction task on a regression dataset (Bike Sharing). The X axis stands for the TDS score decile, the y axis is the coverage percentage. TDS-Mondrian scores best conditional coverage.}
    \label{fig:cp_bike_sharing}
\end{figure}

Figure \ref{fig:cp_bike_sharing} compares the coverage curves of the four conformal prediction methods on the Bike Sharing dataset on a regression model with 100 estimators. TDS exhibits a near flat trend slope across all error deciles and in particular on the most difficult samples.

\subsubsection{Ablation}
As a design policy, most of the method parameters are configurable, as we demonstrated above. While for some tasks, better performance is achieved by removing features (e.g. by zeroing their weight/width), these features contribute significantly on other datasets or tasks. As we reported in this section, in general we found that the default settings produce sufficient results across datasets and tasks and HPO was required only for the minority of the experiments.

We performed two sets of ablation studies to evaluate the effectiveness of trajectory features and the trajectory combinations.

\paragraph{Trajectory features}
We tested robustness by removing the head/tail segments and the signed area descriptor from the trajectory features (ablation=True). 
Table \ref{tab:ablation_deltas} consolidates mean Spearman and Pearson deltas of across four representative regression and classification datasets (adult, WiDS, Bike Sharing, Cal. Housing). The change in ranking quality was small for rank correlation but larger for linear correlation: mean $\Delta Spearman ~ \rho$ = $-1.8$\% and mean $\Delta Pearson ~ r = -8.2\%$ averaged across estimators on regression datasets and mean $\Delta Spearman ~ \rho = -0.21$\%, mean $\Delta Pearson ~ r = +0.52\%$ (full per-configuration results in supplemental material). The per-dataset breakdown shows $Spearman ~ \rho$ fluctuations mostly within 1-2\%, indicating that (i) the trajectory descriptors may be partly redundant for some datasets and (ii) the default combination is a stable choice. Pearson is more sensitive to this ablation on average, but our primary ranking metric Spearman remains stable. We therefore keep the full descriptor set for general use, while noting that simpler variants may suffice on specific datasets without performance loss.

\begin{table}[t]
\centering
\caption{Ablation deltas (ablation - default), reported as percentage change relative to the default. Values are means with 95\% CI over datasets. Table is grouped by
number of estimators in the model. Bold indicates $|\Delta|\ge 2\%$.}
\label{tab:ablation_deltas}

\input{Tables/table-ablation}

\end{table}

\paragraph{Trajectory combination}
We also compared AL results on a 1000 estimators regression model when restricted to the tree contributions trajectory while omitting the cumulative loss trajectory. As we reported above for the Russian Housing the dataset, this actually improved TDS performance both on AULC and RMSE. We relate it to the noisy labels of this dataset that results in a noisy prediction error trajectory. For the Bike Sharing and Cal. Housing datasets we found a drop of up to 2\% in AULC.

\section{Conclusions and Future Work}
This paper introduced a Trajectory-based Difficulty Score (TDS) that quantifies sample-level difficulty in gradient boosting ensembles by analyzing prediction dynamics across boosting rounds. Grounded in boosting theory, TDS captures how margin evolution and functional gradient descent convergence correlate with prediction error, reflecting both epistemic and aleatoric uncertainties. TDS serves as a unified difficulty signal, bridging model diagnostics to downstream tasks while the base boosting model remains unchanged, and TDS integrates into existing pipelines.
Experimental validation across diverse benchmarks demonstrates consistent improvements in data efficiency, risk-coverage calibration, and conditional coverage. While challenges remain, especially regarding random label noise and classification early-stage performance, the theoretical grounding and empirical results position TDS as a valuable tool for enhancing reliability, interpretability, and efficiency in gradient boosting applications.

\subsection{Key Contributions}
We demonstrated that trajectory features: variance, sign switches, peak magnitudes, and head-tail segment statistics yield strong rank correlation with prediction error. \\
TDS-Segment clusters high-difficulty samples by SHAP representations, supporting failure mode discovery and exposing interpretable sub-populations (e.g., specific geographic regions on California Housing, age and hospitalization patterns on WiDS ICU Survival).
TDS serves as the decision mechanism for a new family of methods across three core tasks:
\begin{enumerate}
    \item In active learning, TDS-guided selection achieved lower AULC and superior final RMSE/LL, with strong late-stage performance on classification at deeper ensembles.
    \item In selective prediction, TDS improved risk-coverage trade-offs, achieving the lowest NAURC on classification. 
    \item For conformal prediction, TDS-Mondrian yielded the flattest conditional coverage across difficulty deciles (lowest MACE and MaxCE). 
\end{enumerate}

\subsection{Limitations}
\begin{itemize}
    \item TDS struggles with randomly mislabeled samples, as trajectories computed from corrupted labels create artificial difficulty signals uncorrelated with true complexity (e.g., Russian Housing Market dataset).
    \item In classification, TDS exhibits slower early-stage active learning convergence when class overlap dominates, potentially selecting redundant boundary samples. TDS introduces hyperparameters (delta, head\_size, tail\_size) requiring dataset-specific optimization in minority cases.
    \item Computational overhead ($\approx$25 seconds for 30 AL iterations) exceeds instantaneous baselines but remains practical.
    \item TDS is currently specifically designed for gradient boosting and does not yet extend to other architectures.
\end{itemize}

\subsection{Future Work}
\begin{enumerate}
    \item Extend TDS to new domains and architectures e.g. vision on convolutional neural networks and language models on transformer architectures.
    \item Explore new applications to benefit from TDS e.g. adversarial training, anomaly and Out Of Distribution (OOD) detection.
\end{enumerate}

\section{Declaration}
\textbf{Funding.} his research received no specific grant from any funding agency in the public, commercial, or not-for-profit sectors.

\textbf{Conflict of interest.} The authors declare that they have no conflict of interest.

\textbf{Ethical approval.} This article does not contain any studies with human participants or animals performed by any of the authors beyond the use of fully de-identified benchmark datasets.

\textbf{Declaration of generative AI and AI-assisted technologies in the manuscript preparation process}
During the preparation of this work the author(s) used Chat GPT in order to assist in language-editing and programming code. After using this tool/service, the author(s) reviewed and edited the content as needed and take(s) full responsibility for the content of the published article.


\bibliography{sn-article}% common bib file

\end{document}

%% file: Tables/table_datasets.tex
\begin{tabular}{lccc}
        \hline
        \textbf{Name} & \textbf{Domain} & \textbf{Features} & \textbf{Rows} \\
         \hline
         \multicolumn{4}{c}{\textbf{Regression}}\\
        \hline
        Bike Sharing \cite{bike_sharing_275} & Transport& 16 & 17,379 \\
        Cal. Housing \cite{scikit-learn} & Business & 8 & 20,640 \\
        CT Slice \cite{relative_location_of_ct_slices_on_axial_axis_206} & Medical & 385 & 53,500\\
        Diabetes \cite{scikit-learn} & Health & 10 & 442 \\
        Protein Tertiary \cite{physicochemical_properties_of_protein_tertiary_structure_265} & Biology & 9 & 45,730 \\
        Russian Housing \cite{sberbank-russian-housing-market} & Business & 291 & 30,471 \\
       Superconductivity \cite{superconductivty_data_464} & Science & 81 & 21,263 \\
        Synthetic & N/A & 15 & 20,000 \\
        Wine Quality \cite{wine_quality_186} & Chemistry & 11 & 6,497 \\
        Year Prediction \cite{year_prediction_msd_203} & Audio & 90 & 515,345 \\
        \hline
        \multicolumn{4}{c}{\textbf{Classification}}\\
        \hline
        Adult \cite{adult_2} & Social & 14 & 48,842 \\
        Banknote Auth. \cite{banknote_authentication_267} & Business & 4 & 1,372 \\
        Bank Marketing \cite{bank_marketing_222} & Business & 16 & 45,211 \\
        Breast Cancer \cite{breast_cancer_wisconsin_(diagnostic)_17} & Health & 30 & 569 \\
        Credit Default \cite{default_of_credit_card_clients_350} & Business & 15& 690\\
        Synthetic & N/A & 15 & 20,000 \\
        Jannis \cite{grinsztajn2022tree} & Other & 54 & 57,580 \\
        Stroke \cite{mxfb} & Health & 12 & 5,110 \\
        WiDS Datathon \cite{widsdatathon2020} & Health & 186 & 91,713 \\
        \hline
    \end{tabular}
    \caption{Overview of datasets with various tasks and domains.}

%% file: Tables/table_difficulty_correlations.tex
\begin{tabular}{c c c}
\hline
  \textbf{Method} & \textbf{Pearson \textbf{$r$}} & \textbf{Spearman \textbf{$\rho$}} \\
\hline
 \multicolumn{3}{c}{\textbf{Classification}}\\
\hline
 \multicolumn{3}{c}{100 trees}\\
\hline
TDS & 0.375 ± 0.092 & 0.904 ± 0.046 \\
DIH & 0.771 ± 0.059 & 0.756 ± 0.067 \\
Class Overlap & 0.742 ± 0.062 & 0.724 ± 0.060 \\
kDN & 0.629 ± 0.096 & 0.549 ± 0.074 \\
\hline
 \multicolumn{3}{c}{300 trees}\\
\hline
TDS & 0.370 ± 0.090 & 0.884 ± 0.044 \\
DIH & 0.770 ± 0.056 & 0.754 ± 0.067 \\
Class Overlap & 0.742 ± 0.060 & 0.722 ± 0.061 \\
kDN & 0.625 ± 0.096 & 0.545 ± 0.075 \\
\hline
 \multicolumn{3}{c}{1000 trees}\\
\hline
TDS & 0.360 ± 0.086 & 0.841 ± 0.035 \\
DIH & 0.767 ± 0.053 & 0.749 ± 0.071 \\
Class Overlap & 0.740 ± 0.055 & 0.716 ± 0.064 \\
kDN & 0.618 ± 0.097 & 0.533 ± 0.078 \\
\hline
 \multicolumn{3}{c}{\textbf{Regression}}\\
\hline
 \multicolumn{3}{c}{100 trees}\\
\hline
RF var & 0.635 ± 0.108 & 0.519 ± 0.073 \\
kDN & 0.544 ± 0.121 & 0.426 ± 0.122 \\
TDS & 0.239 ± 0.056 & 0.425 ± 0.099 \\
QRF PI width & 0.422 ± 0.062 & 0.420 ± 0.096 \\
Cook distance & 0.193 ± 0.112 & 0.374 ± 0.143 \\
\hline
 \multicolumn{3}{c}{300 trees}\\
\hline
RF var & 0.567 ± 0.102 & 0.483 ± 0.091 \\
QRF PI width & 0.382 ± 0.059 & 0.404 ± 0.113 \\
TDS & 0.168 ± 0.092 & 0.393 ± 0.172 \\
kDN & 0.466 ± 0.121 & 0.364 ± 0.122 \\
Cook distance & 0.176 ± 0.120 & 0.290 ± 0.119 \\
\hline
 \multicolumn{3}{c}{1000 trees}\\
\hline
RF var & 0.540 ± 0.092 & 0.466 ± 0.094 \\
QRF PI width & 0.368 ± 0.056 & 0.397 ± 0.115 \\
kDN & 0.436 ± 0.114 & 0.334 ± 0.112 \\
TDS & 0.106 ± 0.103 & 0.317 ± 0.174 \\
Cook distance & 0.171 ± 0.118 & 0.259 ± 0.110 \\
\hline
\end{tabular}

%% file: Tables/table_al_summary.tex
\begin{tabular}{c c c c}
\hline
\textbf{Method} & \textbf{AULC↓} & \textbf{RMSE (Val)} & \textbf{RMSE (Test)} \\
\hline
\multicolumn{4}{c}{\textbf{100 Estimators}}\\
\hline
 \underline{TDS} & \textbf{13.093} & \textbf{12.098} & \textbf{13.491} \\
 Query by Comm. & 13.448 & 12.239 & 13.627 \\
 \underline{TDS-Segment} & 13.899 & 12.635 & 14.095 \\
 Uncertainty-Res. & 13.983 & 12.492 & 14.029 \\
 Core Set & 13.939 & 12.617 & 14.054 \\
 BatchBALD & 14.011 & 12.735 & 13.872 \\
 Learning Loss & 14.078 & 12.652 & 14.028 \\
 Random & 14.143 & 12.663 & 14.164 \\
 GLISTER & 14.187 & 12.714 & 14.161 \\
 BADGE & 14.405 & 12.887 & 14.337 \\
 Uncertainty-Var. & 14.598 & 13.385 & 14.611 \\
 \hline
\multicolumn{4}{c}{\textbf{300 Estimators}}\\
\hline
 \underline{TDS} & \textbf{12.369} & \textbf{11.127} & \textbf{12.427} \\
 Query by Comm. & 12.622 & 11.274 & 12.594 \\
 Core Set & 12.965 & 11.506 & 12.940 \\
 BatchBALD & 12.985 & 11.589 & 12.974 \\
 \underline{TDS-Segment} & 13.007 & 11.675 & 12.888 \\
 Random & 13.147 & 11.536 & 13.022 \\
 GLISTER & 13.323 & 11.801 & 13.174 \\
 Learning Loss & 13.413 & 11.769 & 13.297 \\
 Uncertainty-Res. & 13.366 & 11.702 & 13.019 \\
 BADGE & 13.463 & 11.784 & 13.336 \\
 Uncertainty-Var. & 13.529 & 11.871 & 13.268 \\
\hline
\multicolumn{4}{c}{\textbf{1000 Estimators}}\\
\hline
 Query by Comm. & \textbf{12.427} & \textbf{11.017} & 12.326 \\
 \underline{TDS} & 12.438 & 11.081 & \textbf{12.312} \\
 BatchBALD & 12.673 & 11.305 & 12.694 \\
 Core Set & 12.757 & 11.246 & 12.668 \\
 Random & 12.955 & 11.280 & 12.759 \\
 GLISTER & 13.163 & 11.369 & 12.906 \\
 Uncertainty-Res. & 13.222 & 11.515 & 12.789 \\
 \underline{TDS-Segment} & 13.266 & 11.733 & 12.955 \\
 Uncertainty-Var. & 13.341 & 11.629 & 13.124 \\
 Learning Loss & 13.405 & 11.600 & 12.995 \\
 BADGE & 13.353 & 11.472 & 12.949 \\
\hline

\end{tabular}

%% file: Tables/table_al_summary_cls.tex
\begin{tabular}{c c c c}
\hline
\textbf{Method} & \textbf{AULC} & \textbf{LL (Val)} & \textbf{LL (Test)↓} \\
\hline
\multicolumn{4}{c}{\textbf{100 Estimators}}\\
\hline
\underline{TDS} & 0.5865& 0.5813 & 0.5896 \\
\underline{TDS-Segment} & 0.5878& 0.5824 & 0.5905 \\
Core Set & 0.5878& 0.5837 & 0.5923 \\
BADGE & 0.5873& 0.5829 & 0.5923 \\
Query By Comm. & 0.5877& 0.5827 & 0.5927 \\
Learning Loss & 0.5875& 0.5832 & 0.5932 \\
random & 0.5882& 0.5830 & 0.5934 \\
Uncert. Entropy & 0.5877& 0.5827 & 0.5936 \\
Exp. Model Change & 0.5878& 0.5827 & 0.5937 \\
Uncert. Least Conf. & 0.5878& 0.5827 & 0.5938 \\
Uncert. Res. & 0.5878& 0.5827 & 0.5938 \\
\hline
\multicolumn{4}{c}{\textbf{300 Estimators}}\\
\hline
\underline{TDS} & 0.5778& 0.5723 & 0.5815 \\
Uncert. Entropy & 0.5786& 0.5741 & 0.5844 \\
Exp. Model Change & 0.5786& 0.5740 & 0.5844 \\
Uncert. Least Conf. & 0.5786& 0.5740 & 0.5845 \\
Uncert. Res. & 0.5786& 0.5740 & 0.5845 \\
BADGE & 0.5786& 0.5737 & 0.5847 \\
Query By Comm. & 0.5786& 0.5733 & 0.5849 \\
Core Set & 0.5789& 0.5748 & 0.5851 \\
random & 0.5794& 0.5742 & 0.5853 \\
Learning Loss & 0.5782& 0.5738 & 0.5855 \\
\underline{TDS-Segment} & 0.5786& 0.5727 & 0.5873 \\
\hline
\multicolumn{4}{c}{\textbf{1000 Estimators}}\\
\hline
\underline{TDS} & 0.5582& 0.5529 & 0.5655 \\
Uncert. Entropy & 0.5593& 0.5545 & 0.5664 \\
Exp. Model Change & 0.5593& 0.5546 & 0.5664 \\
Learning Loss & 0.5585& 0.5527 & 0.5665 \\
Uncert. Least Conf. & 0.5590& 0.5545 & 0.5667 \\
Uncert. Res. & 0.5590& 0.5545 & 0.5667 \\
Query By Comm. & 0.5599& 0.5544 & 0.5668 \\
random & 0.5610& 0.5549 & 0.5672 \\
Core Set & 0.5599& 0.5545 & 0.5674 \\
BADGE & 0.5589& 0.5533 & 0.5680 \\
\underline{TDS-Segment} & 0.5600& 0.5537 & 0.5724 \\
\hline

\end{tabular}

%% file: Tables/table_selection_summary.tex
\begin{tabular}{c c c }
\hline
\textbf{Estimators} & \textbf{Method}& \textbf{NAURC}\\
\hline
\multicolumn{3}{c}{\textbf{Regression}}\\
\hline
100 & \textbf{Ensemble Var.} & \textbf{0.319} \\
& QRF Width & 0.484 \\
& TDS & 0.486 \\
300 & \textbf{Ensemble Var.} & \textbf{0.343} \\
& QRF Width & 0.496 \\
& TDS & 0.404 \\
1000 & Ensemble Var. & 0.351 \\
& QRF Width & 0.510 \\
& \textbf{TDS} & \textbf{0.291} \\
 \hline
\multicolumn{3}{c}{\textbf{Classification}}\\
\hline 
100 & Entropy & 0.725 \\
& Margin & 0.725 \\
& Max Softmax & 0.725 \\
& \textbf{TDS} & \textbf{0.716} \\
300 & Entropy & 0.725 \\
& Margin & 0.725 \\
& Max Softmax & 0.725 \\
& \textbf{TDS} & \textbf{0.716} \\
1000 & Entropy & 0.711 \\
& Margin & 0.711 \\
& Max Softmax & 0.711 \\
& \textbf{TDS} & \textbf{0.668} \\
\hline
\end{tabular}

%% file: Tables/table_cp_summary.tex
\begin{tabular}{c c c c c c }
\hline
\textbf{Est.}& \textbf{Method} & \textbf{Cov.}& \textbf{Width} & \textbf{MACE} & \textbf{MaxCE↓} \\
\hline
\multicolumn{6}{c}{\textbf{Regression}}\\
\hline
100 & TDS & 0.904 & 43.551 & 0.005 & 0.027 \\
& CQR  & 0.904 & 68.140 & 0.019 & 0.087 \\
& Mondrian & 0.905 & 43.963 & 0.017 & 0.127 \\
& vanilla & 0.902 & 45.203 & 0.035 & 0.205 \\
300 & TDS & 0.904 & 42.536 & 0.005 & 0.031 \\
& CQR  & 0.902 & 72.625 & 0.016 & 0.055 \\
& Mondrian & 0.905 & 42.631 & 0.016 & 0.105 \\
& vanilla & 0.902 & 43.261 & 0.032 & 0.170 \\
1000 & TDS & 0.905 & 41.812 & 0.004 & 0.025 \\
& CQR & 0.902 & 72.625 & 0.016 & 0.059 \\
& Mondrian & 0.905 & 41.789 & 0.014 & 0.081 \\
& vanilla & 0.901 & 42.384 & 0.028 & 0.136 \\
\hline
\multicolumn{6}{c}{\textbf{Classification}}\\
\hline
100 & TDS & 0.902 & 1.102 & 0.008 & 0.036 \\
& Mondrian & 0.904 & 1.111 & 0.008 & 0.042 \\
& vanilla & 0.901 & 1.031 & 0.049 & 0.339 \\
300 & TDS & 0.902 & 1.102 & 0.008 & 0.036 \\
& Mondrian & 0.904 & 1.111 & 0.008 & 0.042 \\
& vanilla & 0.901 & 1.031 & 0.049 & 0.339 \\
1000 & TDS & 0.904 & 1.147 & 0.004 & 0.023 \\
& Mondrian & 0.906 & 1.161 & 0.007 & 0.030 \\
& vanilla & 0.901 & 1.052 & 0.062 & 0.330 \\
\hline
\end{tabular}

%% file: Tables/table-ablation.tex
\begin{tabular}{c c c c}
\toprule
\textbf{Dataset} & \textbf{Estimators} & $\mathbf{\Delta\rho}$  & $\mathbf{\Delta r}$ \\
\midrule
 \multicolumn{4}{c}{\textbf{Regression}}\\
\midrule
  bike\_sharing & 100 & -0.2\% & \textbf{-3.0\%} \\
& 1000 & -1.1\% & \textbf{-5.3\%} \\
california\_housing & 100 & +0.6\% & \textbf{-6.2\%} \\
& 1000 & \textbf{-6.3\%} & \textbf{-18.5\%} \\
\textit{Mean} &  & -1.8\% & \textbf{-8.2\%} \\
\midrule
\multicolumn{4}{c}{\textbf{Classification}}\\
\midrule
  adult& 100 & +0.26\%& -0.07\%\\
& 1000 & +0.12\%& +1.7\%\\
WiDS& 100 & -0.8\%& +1.37\%\\
& 1000 & -0.43\%& -0.91\%\\
\textit{Mean} &  & -0.21\%& +0.52\%\\
\bottomrule
\end{tabular}